\documentstyle[aaai]{article}
\newlength{\axiomwidth}
\newtheorem{lemm}{Lemma}
\newtheorem{defn}{Definition}
\newtheorem{thm}{Theorem}
\newtheorem{expl}{Example}
\newtheorem{prop}{Proposition}

\newenvironment{definition}{ \begin{defn} \em }{ {\hfill $\Box$} \end{defn} }

\newenvironment{example}{ \begin{expl} \em }{ {\hfill $\Box$} \end{expl} }

\newcommand{\NI}{\noindent}

\newcommand{\PR}{\mbox{$\ \vdash\ $}}

\newcommand{\E}{\cal E}
\newcommand{\Take}{\mbox{\textit{Take}}}
\newcommand{\Picture}{\mbox{\textit{Picture}}}
\newcommand{\Loaded}{\mbox{\textit{Loaded}}}
\newcommand{\Digital}{\mbox{\textit{Digital}}}
\newcommand{\TurnOn}{\mbox{\textit{TurnOn}}}
\newcommand{\TurnOff}{\mbox{\textit{TurnOff}}}
\newcommand{\Running}{\mbox{\textit{Running}}}
\newcommand{\Petrol}{\mbox{\textit{Petrol}}}
\newcommand{\Empty}{\mbox{\textit{Empty}}}
\newcommand{\JumpStart}{\mbox{\textit{JumpStart}}}

\newcommand{\InjectA}{\mbox{\textit{InjectA}}}
\newcommand{\InjectB}{\mbox{\textit{InjectB}}}
\newcommand{\InjectC}{\mbox{\textit{InjectC}}}
\newcommand{\InjectD}{\mbox{\textit{InjectD}}}
\newcommand{\InjectE}{\mbox{\textit{InjectE}}}
\newcommand{\Infected}{\mbox{\textit{Infected}}}
\newcommand{\Protected}{\mbox{\textit{Protected}}}
\newcommand{\TypeO}{\mbox{\textit{TypeO}}}
\newcommand{\TypeA}{\mbox{\textit{TypeA}}}
\newcommand{\TypeB}{\mbox{\textit{TypeB}}}
\newcommand{\Weak}{\mbox{\textit{Weak}}}
\newcommand{\Strong}{\mbox{\textit{Strong}}}
\newcommand{\Bite}{\mbox{\textit{Bite}}}
\newcommand{\Expose}{\mbox{\textit{Expose}}}
\newcommand{\Danger}{\mbox{\textit{Danger}}}
\newcommand{\Allergic}{\mbox{\textit{Allergic}}}

\newcommand{\Initiation}{\mbox{\textit{Initiation}}}
\newcommand{\HoldsAt}{\mbox{\textit{HoldsAt}}}
\newcommand{\Termination}{\mbox{\textit{Termination}}}
\newcommand{\HappensAt}{\mbox{\textit{HappensAt}}}

\newcommand{\PGe}{\mbox{\textit{PG}}}
\newcommand{\NGe}{\mbox{\textit{NG}}}
\newcommand{\PA}{\mbox{\textit{PA}}}
\newcommand{\NA}{\mbox{\textit{NA}}}
\newcommand{\PP}{\mbox{\textit{PP}}}
\newcommand{\NP}{\mbox{\textit{NP}}}

\newcommand{\ARE}{\mathcal{A}_{\E}}
\newcommand{\PED}{P_{\E}(D)}

\begin{document}

\title{$\E$-RES - A System for Reasoning about Actions,
Events and Observations 
}

\author{ {\bf Antonis Kakas} \\
University of Cyprus,\\
Nicosia, CYPRUS\\
{\it antonis@cs.ucy.ac.cy} \\
\And
{\bf Rob Miller}  \\
University College London, \\
London, U.K. \\
{\it rsm@ucl.ac.uk} \\
\And
{\bf Francesca Toni}  \\
Imperial College, \\
London, U.K. \\
{\it ft@doc.ic.ac.uk} \\
}

%

\maketitle


\begin{abstract}
\noindent
$\E$-RES is a system that implements the Language $\E$, a
logic for reasoning about narratives of action occurrences and
observations. $\E$'s semantics is model-theoretic, but this
implementation is based on a sound and complete reformulation of $\E$
in terms of argumentation, and uses general computational
techniques of argumentation frameworks. The system
derives sceptical non-monotonic consequences of a given
reformulated theory which exactly correspond to consequences
entailed by $\E$'s model-theory.
The computation relies on a complimentary
ability of the system to derive credulous non-monotonic
consequences together with a set of supporting assumptions
which is sufficient for the (credulous) conclusion to hold.
$\E$-RES
allows theories to contain general action laws, statements about
action occurrences,
observations and statements of ramifications (or universal laws).
It is able to derive consequences both forward
and backward in time. This paper gives a short overview
of the theoretical basis of $\E$-RES and illustrates
its use on a variety of examples. Currently,
$\E$-RES is being extended so that the system can be used
for planning.
\end{abstract}

%
%
%
%
%
%
%
%
%
%
%
%
%
%
%
%
%
%
%
%
%
%
%
%
%
%
%
%
%
%
%
%
%
%
%
%
%
%
%
%
%
%

\section{General Information}\label{GeneralInfoSection}


$\E$-RES is a system for modeling and reasoning about
dynamic systems. Specifically, it implements the Language $\E$ \cite{KaMi97,KaMi97bis}, a
specialist logic for reasoning about narratives of action occurrences and
observations. $\E$-RES is implemented in SICStus Prolog and runs
on any platform for which SICStus is supported
(e.g.\ Windows, Linux, UNIX, Mac).
The program is about 300 lines long (a URL
is given at the end of the paper). The semantics of $\E$
is model-theoretic, but this
implementation is based on a sound and complete reformulation of $\E$
in terms of argumentation, and uses general computational
techniques of argumentation frameworks.
To describe the operation and utility of $\E$-RES, it is
necessary to first review the Language $\E$.


\subsection{The Language $\E$}\label{LanguageESection}

Like many logics, the Language $\E$ is really a collection of languages, since the
particular vocabulary employed depends on the domain being modeled.
The domain-dependent vocabulary always consists of a set of
{\it fluent constants}, a set of {\it  action constants},
and a partially ordered set
$\langle \Pi , \preceq \rangle$ of {\it time-points}.
A {\it fluent literal} is either a fluent constant $F$
or its negation $\neg F$. In the current implementation of
$\E$-RES the only time structure that is supported is that
of the natural numbers, so we restrict our attention here to domains
of this type, using the standard ordering relation $\leq$ in all
examples.

{\em Domain descriptions} in the Language ${\cal E}$ are collections
of four kinds of statements
(where $A$ is an action constant, $T$ is a time-point,
$F$ is a fluent constant, $L$ is a fluent literal
and  $C$ is a set of
fluent literals):

\begin{itemize}
    \item  {\em t-propositions} (``t'' for ``time-point''),
           of the form
           \vspace{-0.2 em}
          \begin{center}
               $L \mbox{ {\tt holds-at} } T$
           \end{center}
           \vspace{-0.2 em}

    \item  {\em h-propositions} (``h'' for ``happens''),
           of the form
           \vspace{-0.2 em}
          \begin{center}
               $A \mbox{ {\tt happens-at} } T$
           \end{center}
           \vspace{-0.2 em}

    \item  {\em c-propositions} (``c'' for ``causes''), either of the form
           \vspace{-0.2 em}
           \begin{center}
               $A \mbox{ {\tt initiates} } F  \mbox{ {\tt when} } C$
           \end{center}
           \vspace{-0.2 em}
            or of the form
           \vspace{-0.2 em}
           \begin{center}
               $A \mbox{ {\tt terminates} } F  \mbox{ {\tt when} } C$
            \end{center}
           \vspace{-0.2 em}

    \item  {\em r-propositions} (``r'' for ``ramification''), of the form
           \vspace{-0.2 em}
           \begin{center}
               $L  \mbox{ {\tt whenever} } C$.
            \end{center}
           \vspace{-0.2 em}
\end{itemize}
The precise semantics of $\E$ is described in \cite{KaMi97} and \cite{KaMi97bis}.
T-propositions are used to record observations that particular
fluents hold or do not hold at particular time-points, and
h-propositions are used to state that particular actions occur at
particular time-points. C-propositions state general ``action
laws'' -- the intended meaning of ``$A$ {\tt initiates} $F$ {\tt when} $C$''
is
``$C$ is a
minimally sufficient set of conditions for an occurrence of $A$ to have
an initiating effect on $F$''. (When $C$ is empty the proposition is
stated simply as ``$A$ {\tt initiates} $F$''.)
R-propositions serve a dual role in that they describe both static
constraints between fluents and ways in which fluents may be
indirectly affected by action occurrences. The intended meaning of
``$L$ {\tt whenever} $C$'' is ``at every
time-point that $C$ holds, $L$ holds, and hence every action
occurrence that brings about $C$ also brings about $L$''.

$\E$'s semantics is perhaps best understood by examples, and so several
are given in the next sub-section. The key features of the
semantics are as follows.

\sloppy
\begin{itemize}
    \vspace{-0.2 em}
    \item  Models are simply mappings of fluent/time-point pairs
    to $\{\mbox{\textit{true}},\mbox{\textit{false}}\}$ which satisfy various
    properties relating to the propositions in the domain.

    \vspace{-0.2 em}
    \item  The semantics describes {\em entailment} ($\models$) of extra t-propositions (but not
    h-, c- or r-propositions) from domain descriptions.

    \vspace{-0.2 em}
    \item  $\E$ is monotonic as regards addition of t-propositions
    to domain descriptions, but non-monotonic (in order to eliminate
    the frame problem) as regards addition of
    h-, c- and r-propositions. The semantics encapsulates the
    assumptions that (i) no actions occur other than those explicitly
    represented by h-propositions, (ii) actions have no
    direct effects other than those explicitly described by
    c-propositions, and (iii) actions have no indirect effects other than
    those that can be explained by ``chains'' of r-propositions in the domain
    description. (Technically,
    these ``chains'' are defined using the notion of a least fixed
    point.)

    \vspace{-0.2 em}
    \item  The semantics ensures that fluents have a \textit{default
    persistence}. In each model, fluents change truth values only at time-points
    (called {\em initiation points} and {\em termination points}) where an
    h-proposition and a c-proposition (whose preconditions are
    satisfied in the model) combine to cause a change, or where an
    h-proposition, a c-proposition and a ``chain'' of r-propositions
    all combine to give an indirect or knock-on effect. All effects
    (direct and indirect) of an
    action occurrence are instantaneous, i.e.\ all changes are
    apparent immediately after the occurrence.

    \vspace{-0.2 em}
    \item  As well as indicating how the effects of action occurrences
    instantaneously propagate, r-propositions place constraints on
    which combinations of t-propositions referring to the same
    time-point are allowable. In this latter respect they behave as ordinary
    classical implications.

\end{itemize}
\fussy


\subsection{Example Language $\E$ Domain Descriptions}\label{EExamplesSection}

Each of the following domain descriptions illustrates how $\E$ supports
particular modes of reasoning about the effects of actions.
These domain descriptions are used in subsequent sections of the
paper to illustrate the functionality of the $\E$-RES system.


\begin{example}\label{VaccinationsExample} {\it (Vaccinations)}\\
This example concerns vaccinations against a particular
disease. Vaccine A only provides protection for people with
blood type O, and vaccine B only works on people with blood
type other than O. Fred's blood type is not
known, so he is injected with vaccine A at 2 o'clock and
vaccine B at 3 o'clock. To describe this scenario we need a
vocabulary of two action constants \InjectA\ and \InjectB, and two
fluent constants \Protected\ and \TypeO. The domain description $D_{v}$
consists of two c-propositions and two h-propositions:\\

\noindent
\begin{tabular*}{\axiomwidth}{l@{\extracolsep{\fill}}r@{}}
{$\InjectA \mbox{ {\tt initiates} } \Protected \mbox{ {\tt when} } \{ \TypeO \}$} & {($D_{v}$1)}\\
{$\InjectB \mbox{ {\tt initiates} } \Protected \mbox{ {\tt when} } \{ \neg \TypeO \}$} & {($D_{v}$2)}\\
{$\InjectA \mbox{ {\tt happens-at} } 2$} & {($D_{v}$3)}\\
{$\InjectB \mbox{ {\tt happens-at} } 3$} & {($D_{v}$4)}\\
{} & {}
\end{tabular*}

\noindent
If we now consider some time later than 3 o'clock, say 6 o'clock,
we can see intuitively that Fred should be protected, and indeed it is
the case that
\begin{center}
$D_{v} \models \Protected \mbox{ {\tt holds-at} } 6$.
\end{center}
This is because there are two classes of models for this domain. In
models of the first type, \TypeO\ holds for all time-points, so that
($D_{v}$1) and ($D_{v}$3) combine to form an initiation point for
\Protected\ at $2$. In models of the second type, \TypeO\ does not hold
for any time-point, and so ($D_{v}$2) and ($D_{v}$4) combine to form an initiation point for
\Protected\ at $3$. In either type of model, \Protected\ then persists
from its initiation point up to time-point $6$, since the fluent has no
intervening termination points to override its initiation. Note,
however, that there are no default assumptions directly attached to
t-propositions, so that for any time $T \leq 3$ it is neither the case
that $D_{v}$ entails $\Protected \mbox{ {\tt holds-at} } T$ nor the
case that $D_{v}$ entails $\neg \Protected \mbox{ {\tt holds-at} } T$.
\end{example}


\begin{example}\label{PhotographsExample} {\it (Photographs)}\\
This example shows that the Language $\E$ can be used
to infer information about what conditions hold at the time of
an action occurrence, given other information about what held at
times before and afterwards. It concerns taking a photograph. There
is a single action \Take, and two fluents \Picture\ (representing that
a photograph has been successfully taken) and \Loaded\ (representing that the
camera is loaded with film).
Suppose that the domain
description $D_{p}$ consists of a single c-proposition,
a single h-proposition and two t-propositions:\\

\noindent
\begin{tabular*}{\axiomwidth}{l@{\extracolsep{\fill}}r@{}}
{$\Take \mbox{ {\tt initiates} } \Picture \mbox{ {\tt when} } \{ \Loaded \}$} & {($D_{p}$1)}\\
{$\Take \mbox{ {\tt happens-at} } 2$} & {($D_{p}$2)}\\
{$\neg \Picture \mbox{ {\tt holds-at} } 1$} & {($D_{p}$3)}\\
{$\Picture \mbox{ {\tt holds-at} } 3$} & {($D_{p}$4)}\\
{} & {}
\end{tabular*}

\noindent
Since a change occurs in the truth value of \Picture\ between $1$ and
$3$, in all models an action must occur at
some time-point between $1$ and $3$ whose initiating
conditions for the property \Picture\ are satisfied at that
point. The only candidate is the \Take\ occurrence at 2,
whose condition for initiating \Picture\ is \Loaded. Hence
\begin{center}
$D_{p} \models \Loaded \mbox{ {\tt holds-at} } 2$.
\end{center}
Indeed, by the persistence of \Loaded\ (in the absence of  possible
initiation or termination points for this fluent),
for any time $T$,
$D_{p} \models \Loaded \mbox{ {\tt holds-at} } T$.
\end{example}


\begin{example}\label{CarsExample} {\it (Cars)}\\
This example illustrates the use of r-propositions, and shows how the
effects of later action occurrences override the effects of earlier action
occurrences. It concerns a car engine. The fluent \Running\ represents
that the engine is running, the fluent \Petrol\ represents that there
is petrol (gas) in the tank, the action \TurnOn\ represents the
action of turning on the engine, the action \TurnOff\ represents the
action of turning off the engine, and the action \Empty\
represents the event of the tank becoming empty (or the action of
someone emptying the tank). We describe a narrative where the engine
is initially running, is turned off at time $2$, is turned back on at
time $5$, and runs out of petrol at time $8$. We also want to state
the general constraint that the engine cannot run without petrol. The
domain description $D_{c}$
consists of:\\

\noindent
\begin{tabular*}{\axiomwidth}{l@{\extracolsep{\fill}}r@{}}
{$\TurnOn \mbox{ {\tt initiates} } \Running \mbox{ {\tt when} } \{ \Petrol \}$} & {($D_{c}$1)}\\
{$\TurnOff \mbox{ {\tt terminates} } \Running$} & {($D_{c}$2)}\\
{$\Empty \mbox{ {\tt terminates} } \Petrol$} & {($D_{c}$3)}\\
{$\neg \Running \mbox{ {\tt whenever} } \{\neg \Petrol \}$} & {($D_{c}$4)}\\
{$\Running \mbox{ {\tt holds-at} } 1$} & {($D_{c}$5)}\\
{$\TurnOff \mbox{ {\tt happens-at} } 2$} & {($D_{c}$6)}\\
{$\TurnOn \mbox{ {\tt happens-at} } 5$} & {($D_{c}$7)}\\
{$\Empty \mbox{ {\tt happens-at} } 8$} & {($D_{c}$8)}\\
{} & {}
\end{tabular*}

\noindent
The Language $\E$ supports the following conclusions concerning the
fluents \Running\ and \Petrol:\\
%
%
(i)
For $T \leq 2$, $D_{c} \models \Running \mbox{ {\tt holds-at} } T$.
This is because of ($D_{c}$5), and because there are no
relevant action occurrences before time $2$ to override \Running's
default persistence.\\
%
%
(ii)
For $T \leq 8$, $D_{c} \models \Petrol \mbox{ {\tt holds-at} } T$.
This is because we obtain $\Petrol \mbox{ {\tt holds-at} }
1$ directly from ($D_{c}$4) and ($D_{c}$5), and because there are no
relevant action occurrences before time $8$ to override \Petrol's
default persistence. Note that in this case ($D_{c}$4) has been used
(in the contrapositive) in its capacity as a static constraint at
time $1$.\\
%
%
(iii)
For  $2 < T \leq 5$, $D_{c} \models \neg \Running \mbox{ {\tt holds-at} } T$.
This is because ($D_{c}$2) and ($D_{c}$6) combine to form a
termination point for \Running\ at $2$ (in all models).\\
%
%
(iv)
For  $5 < T \leq 8$, $D_{c} \models \Running \mbox{ {\tt holds-at} } T$.
This is because ($D_{c}$1) and ($D_{c}$7) combine to form an
initiation point for \Running\ at $5$, and this
overrides the earlier termination point (for all times greater than
$5$).\\
%
%
(v)
For $T > 8$, $D_{c} \models \neg \Petrol \mbox{ {\tt holds-at} } T$.
This is because ($D_{c}$3) and ($D_{c}$8) combine to form a
termination point for \Petrol\ at $8$.\\
%
%
(vi)
For  $T > 8$, $D_{c} \models \neg \Running \mbox{ {\tt holds-at} } T$.
This is because ($D_{c}$3), ($D_{c}$4) and ($D_{c}$8) combine to form a
termination point for \Running\ at $8$ which overrides the earlier
initiation point.
%

Note that $\E$ does not allow r-propositions to be
used in the contrapositive to generate extra initiation or
termination points. For example, if we were to add the two
propositions\\

\noindent
\begin{tabular*}{\axiomwidth}{l@{\extracolsep{\fill}}r@{}}
{$\JumpStart \mbox{ {\tt initiates} } \Running$} & {($D_{c}$9)}\\
{$\JumpStart \mbox{ {\tt happens-at} } 11$} & {($D_{c}$10)}\\
{} & {}
\end{tabular*}

\noindent
to the domain description, we would have inconsistency. The
combination of ($D_{c}$9) and ($D_{c}$10) would give an initiation point
for \Running\ at time $11$, so that at subsequent times \Running\
would be true. However, (v) above shows that for such times \Petrol\ is false,
and this contradicts ($D_{c}$4) in its capacity as a static
constraint. ($D_{c}$4) cannot be used in the contrapositive
to ``fix'' this by generating a termination point
for \Petrol\ from the termination point for \Running.
\end{example}

%
%
%
%
%
%
%
%
%
%
%
%
%
%
%
%
%
%
%
%
%
%
%
%
%
%
%
%
%
%
%
%
%
%
%
%
%
%
%
%
%
%

\section{Description of the System}\label{DescriptionSection}

The system relies upon a reformulation
of the Language ${\cal E}$ into argumentation as described in \cite{lpnmr99}.


\subsection{Argumentation Formulation of ${\cal E}$}\label{ArgumentationSection}

A domain description $D$ without t-propositions and without r-propositions is translated
into an {\em argumentation program} $\PED = (B(D), \ARE,$ $\ARE', <_{\E})$,
where $B(D)$ is the
{\em background theory},
$\ARE$ is the {\em argumentation theory}, i.e. a set of {\em argument rules},
$\ARE'\subseteq \ARE$ is the {\em argument base}, and
$<_{\E}$ is a {\em priority relation} over the (ground instances of the) argument rules.
Intuitively, the sentences in the monotonic
background theory can be seen as non-defeasible
argument rules which must belong to any non-monotonic extension of the theory.
These extensions are given by the {\em admissible} subsets of $\ARE'$,
namely subsets that are
%
%
both
{\em non-self-attacking} and
%
%
{\em (counter)attack} any set of argument rules {\em attacking} them.
%
%
Whereas an admissible set can consist only of
argument rules in the argument base,
{\em attacks} against an admissible set
are allowed to be subsets of the larger argument theory.
The exact definition of an attack, which is dependent on the priority
relation $<_{\cal E}$ and the derivation of complimentary literals, is
given in \cite{lpnmr99}.

Both $B(D)$ and $\ARE$ use the predicates $\HappensAt$, $\HoldsAt$,
$\Initiation$ and $\Termination$. $B(D)$ is a set of Horn clauses
corresponding to the h- and c-propositions in $D$ defining the
above predicates expect $\HoldsAt$.
$\ARE$ is a domain independent set of {\em generation}, {\em persistence} and
{\em assumption} rules for $\HoldsAt$. For example, a generation rule is
given by
$\HoldsAt(f,t_{2}) \!\leftarrow \! \Initiation(f,t_{1}),\\
t_{1} \! \prec \! t_{2}$,
where $f$ is any fluent and $t_{1}, t_{2}$ are any two time points.
The relation $<_{\cal E}$ is such that the effects of later events
take priority over the effects of earlier ones (see \cite{lpnmr99}).
Given the translation, results in \cite{lpnmr99} show that
there is  a one-to-one correspondence between
(i)
models of $D$
and maximal admissible sets of arguments of $\PED$, and
(ii)
t-propositions entailed by $D$ and sceptical non-monotonic
consequences of the form $(\neg) \HoldsAt(F,T)$ of $\PED$,
where
a given literal $\sigma$ is a {\em sceptical} (resp.\ {\em credulous})
non-monotonic consequence of  an
argumentation program
iff $B(D) \cup \Delta \vdash \sigma $ for {\em all} (resp.\ {\em some})
maximal admissible extension(s) $\Delta$ of the program.

This method can be applied directly for
conjunctions of literals rather than individual literals.
Hence
the above techniques can be straightforwardly applied
to domains with t-propositions
simply by adding all t-propositions in the domain to the conjunctions of literals
whose entailment we want to check.
Similarly, the above techniques can be directly adapted for
domains with  r-propositions by conjoining to the given literals the conclusion of
ramification statements that are ``fired''.


\subsection{Proof theory}\label{ProofTheorySection}

Given the translation of an $\E$ domain description $D$
into $\PED$, a proof theory 
can be developed directly \cite{lpnmr99},
in terms of {\em derivations of trees},
whose nodes are sets of arguments in $\ARE$ attacking the
arguments in their parent nodes.
Suppose that we wish to demonstrate that a t-proposition
$(\neg) F \mbox{ {\tt holds-at} } T$ is entailed by $D$.
Let
$S_0$ be a (non-self-attacking) set of arguments in $\ARE'$ 
such that $B(D) \cup S_0 \PR (\neg) H \! oldsAt(F,T)$
($S_0$ can be easily built by backward reasoning). 
Two kinds of derivations are defined:
\vspace{-0.1 em}
\begin{itemize}
\item[-]
{\em successful derivations}, building, from
a tree consisting only of the root $S_0$,
a 
tree whose root $S$ is an admissible subset of 
$\ARE'$ such that $S \supseteq S_0$, and
\vspace{-0.1 em}
\item[-]
{\em finitely failed derivations},
guaranteeing the absence of any admissible set of arguments
containing $ S_0$.
\end{itemize}
\vspace{-0.1 em}
Hence the given t-proposition is entailed by $D$ if there exists a successful
derivation
with initial tree consisting only of the root $S_0$ and,
for every set $S_0'$ of argument rules in $\ARE'$ such that
$B(D) \cup S_0'$ derives (via $\vdash$) the complement of the
(literal translation of the) t-proposition,
every derivation for $S_0'$ is finitely failed.


\subsection{Implementation}\label{ImplementationSection}

The system is an implementation of the proof theory presented in \cite{lpnmr99},
but it does not rely explicitly on tree-derivations.
Instead,
it implicitly manipulates trees via their frontiers,
in a way similar to the proof procedure for computing partial stable models of logic programs
in \cite{bob-kave,KM-PRICAI90}.
(See also \cite{KaTo99} for a general discussion of this technique.)

$\E$-RES defines the Prolog predicates
{\tt sceptical/1} and
{\tt credulous/1}.
For some given {\tt Goal} which is a list of literals, with each literal either of the form
{\tt holds(f,t)} or {\tt neg(holds(f,t))} (where the Prolog constant symbols
{\tt f} and  {\tt t}
represent a ground fluent constant $F$ and time point $T$
respectively),
\vspace{-0.1 em}

\begin{itemize}
\item
if {\tt sceptical(Goal)}
succeeds then each literal in {\tt Goal} is a
sceptical non-monotonic consequence of the domain
\vspace{-0.1 em}
\item
if {\tt sceptical(Goal)} finitely fails then
some literal in {\tt Goal} is not a sceptical non-monotonic consequence of the domain
\vspace{-0.1 em}
\item
if {\tt credulous(Goal)}
succeeds then each literal in {\tt Goal} is a
credulous non-monotonic consequence of the domain
\vspace{-0.1 em}
\item
if  {\tt credulous(Goal)} finitely fails then
some literal in {\tt Goal} is not a credulous non-monotonic consequence of the domain.
\end{itemize}
\vspace{-0.1 em}

The implementation also defines the Prolog predicate
{\tt credulous/2}.
This is such that for some given {\tt Goal},
\vspace{-0.1 em}
\begin{itemize}
\item
if {\tt credulous(Goal,X)} succeeds then
each literal in {\tt Goal} is a
credulous non-monotonic consequence of the domain,
and the set of arguments in {\tt X}
provides the corresponding admissible extension of the argumentation program
translation of the domain.
\end{itemize}
\vspace{-0.1 em}


\noindent
Hence {\tt credulous(Goal,X)} can be used to provide an {\em
explanation} {\tt X} for the goal {\tt Goal}.

Domain descriptions in $\E$ may sometimes be described using
meta-level quantification, and $\E$-RES can support
a restricted form of non-propositional programs where
all c-propositions are ``strongly range-restricted'', i.e.\
only of the form $A(\overline{Y}) \mbox{ {\tt initiates} } F(\overline{X})
  \mbox{ {\tt when} } C(\overline{Z})$
where $ \overline{Z} \subseteq \overline{X} \cup \overline{Y}$.
(We assume the usual convention of universal quantification
over the whole proposition.)
However, all h-propositions and queries must be
ground.
Ramifications could also be specified with variables provided
that they all have the general form\\
$L(\overline{Z}) \mbox{ {\tt whenever} } C(\overline{Z_{1}})$
where $ \overline{Z_{1}} \subseteq \overline{Z}$.
However, in the present implementation such statements need to be ground
before they can be handled by the system.


%
%
%
%
%
%
%
%
%
%
%
%
%
%
%
%
%
%
%
%
%
%
%
%
%
%
%
%
%
%
%
%
%
%
%
%
%
%
%
%
%
%

\section{Applying the System}\label{ApplySection}


\subsection{Methodology}\label{MethodologySection}

The system relies upon the
formulation of problems as domains in the Language $\E$,
and a simple and straightforward translation of these $\E$-domains into their
logic-programming based counterparts, which are directly manipulated by the system.
At the time of writing this report,
the translation needs to be performed by hand by the user.
However, the problem of automating this translation presents no conceptual
difficulties, and is scheduled to be implemented in the near future.
As an illustration, consider
Example~\ref{CarsExample}. Its
translation is:\\

\noindent
\begin{tabular*}{\axiomwidth}{l@{\extracolsep{\fill}}r@{}}
{\texttt{initiation(running,T):- }} & {}\\
{\hspace{1.6 em}\texttt{happens(turnOn,T), holds(petrol,T), true.}} & {}\\

{\texttt{termination(running,T):-}} & {}\\
{\hspace{1.6 em}\texttt{happens(turnOff,T), true.}} & {}\\

{\texttt{termination(petrol,T):-}} & {}\\
{\hspace{1.6 em}\texttt{happens(empty,T), true.}} & {}\\

{\texttt{ram(neg(holds(running,T))):-}} & {}\\
{\hspace{1.6 em}\texttt{neg(holds(petrol,T)).}} & {}\\

{\texttt{tprop(holds(running,1)).}} & {}\\

{\texttt{happens(turnOff,2).}} & {}\\

{\texttt{happens(turnOn,5).}} & {}\\

{\texttt{happens(empty,8).}} & {}\\

{} & {}
\end{tabular*}
\vspace{-1 em}

\subsection{Specifics}\label{SpecificsSection}


The system relies upon a logic-based representation of concrete domains.
The system has been developed systematically from its specification
given by the model-theoretic semantics, and
this guarantees its correctness.

The system performs the kind of reasoning which forms the basis of a number of
applications in computer science and artificial intelligence, such as
simulation, fault diagnosis, planning and cognitive robotics.
We are currently studying
extensions of the system that can be used directly to perform planning
in domains that are partially unknown
\cite{nmr00}.


\subsection{Users and Usability}\label{UsersSection}

The use of $\E$-RES requires
knowledge of the Language $\E$, which (like the Language ${\cal A}$ \cite{Gelf93})
has been designed as a high-level
specification tool, in $\E$'s case for modeling dynamic systems as narratives
of action occurrences and observations, where actions can have both
direct and indirect effects. As mentioned above, $\E$-RES is at an
early stage of development, but we aim to soon have a user
interface that will allow domain descriptions to be described
directly in $\E$'s syntax.

%
%
%
%
%
%
%
%
%
%
%
%
%
%
%
%
%
%
%
%
%
%
%
%
%
%
%
%
%
%
%
%
%
%
%
%
%
%
%
%
%

\section{Evaluating the System}\label{EvaluationSection}

The $\E$-RES system is an initial prototype. The prototype has been
evaluated in two distinct ways. First, theoretical results have been
developed (documented in \cite{lpnmr99}) which verify that the system
meets its specification, i.e.\ that it faithfully captures the
entailment relation of the Language $\E$. Second, the system has been
evaluated by testing it on a suite of examples that involve different
modes of reasoning about actions and change.
These examples, which include those given above,
provide ``proof-of-principle''
evidence for the Language $\E$
(and the argumentation approach taken in providing a computational counterpart to it)
as a suitable framework for reasoning about actions and change.

$\E$-RES correctly computes all the t-propositions entailed
by Examples~\ref{VaccinationsExample}, ~\ref{PhotographsExample}
and~\ref{CarsExample}.
This involves reasoning
with incomplete information, reasoning from effects to causes,
reasoning backwards and forwards in time, reasoning about alternative
causes of effects, reasoning about indirect effects, and combining
these forms of reasoning. In the remainder of this section we consider in detail
how $\E$-RES processes a small selection of the queries that can be associated with
these example domains.\\

\NI
\textbf{Testing with Example~\ref{VaccinationsExample}}\\
Example~\ref{VaccinationsExample} can be used to test how $\E$-RES
deals with incomplete information about fluents, and how it is able
to reason with alternatives.
%
%
%
As explained previously, up until time $3$ the
truth value of \Protected\ is unknown. In other words, for times less
than or equal to $3$, the literals \Protected\ and $\neg\Protected$
both hold credulously, but neither holds sceptically. Reflecting
this, for all $t \leq 3$, $\E$-RES succeeds on
\begin{center}
    {\tt credulous([holds(protected,t)])}\\
    {\tt credulous([neg(holds(protected,t))])}
\end{center}
but fails on
\begin{center}
    {\tt sceptical([holds(protected,t)])} \\
    {\tt sceptical([neg(holds(protected,t))])}
\end{center}

After time 3, however, the fluent \Protected\ holds sceptically and
so for all $t < 3 $ the system correctly
succeeds on
\begin{center}
    {\tt sceptical([holds(protected,t)])}
\end{center}
and fails on
\begin{center}
    {\tt sceptical([neg(holds(protected,t))])}.
\end{center}

\NI
\textbf{Testing with Example~\ref{PhotographsExample}}\\
Example~\ref{PhotographsExample} can be used to test how $\E$-RES
can reason from effects to causes, and how it is able
to reason both forwards and backwards in time.
%
%
The observed value of \Picture\ at time $3$ is explained by the fact that
at the time $2$, when a \Take\ action occurred, \Loaded\ held (there is no
alternative way to explain this in the given domain).
Hence the system reasons both backwards and forwards in time so that
\begin{center}
    {\tt sceptical([holds(loaded,2)])}
\end{center}
succeeds. Furthermore, by  persistence,
\begin{center}
    {\tt sceptical([holds(loaded,t)])}
\end{center}
also succeeds for any time {\tt t}.

Note however that if we remove ($D_{p}$3) from the domain,
then \Loaded\ holds only credulously at any time {\tt t}, and hence
the system fails on
\begin{center}
    {\tt sceptical([holds(loaded,t)])}
\end{center}
but succeeds on
\begin{center}
    {\tt credulous([holds(loaded,t)])}.
\end{center}

\NI
If $D_{p}$ is augmented with the
c-proposition\\

\noindent
\begin{tabular*}{\axiomwidth}{l@{\extracolsep{\fill}}r@{}}
{$\Take \mbox{ {\tt initiates} } \Picture \mbox{ {\tt when} } \{ \Digital \}$} & {($D_{p}$5)}\\
{} & {}
\end{tabular*}

\NI
then $\Loaded$ no longer holds sceptically at any time,
since there is now an alternative assumption to explain the observation given by ($D_{p}$4),
namely that \Digital\ holds at $2$. Thus, for any time {\tt t} the
system fails on
\begin{center}
    {\tt sceptical([holds(loaded,t)])}\\
    {\tt sceptical([holds(digital,t)])}
\end{center}
but succeeds on
\begin{center}
    {\tt credulous([holds(loaded,t)])}\\
    {\tt credulous([holds(digital,t)])}.
\end{center}


\noindent
If we pose the query
\begin{center}
    {\tt credulous([holds(picture,3)],X)}.
\end{center}
\noindent
then we will get the two explanations for this observation in terms
of the possible generation rules for $Picture$ and assumptions on
corresponding fluents in their preconditions. These
will be given by the system as:\\
    {\tt X = [rule(gen,picture,3,2),rule(ass,loaded,2)]},\\
    {\tt X = [rule(gen,picture,3,2),rule(ass,digital,2)]}.\\


\NI
\textbf{Testing with Example~\ref{CarsExample}}\\
Example~\ref{CarsExample} can be used to test how $\E$-RES can
reason with ramification statements (r-propositions) and how it can
reason with a series of action occurrences where later occurrences
override the effects of earlier ones.
%
%
As required, $\E$-RES succeeds on each of the following queries:
\begin{center}
    {\tt sceptical([holds(running,0)])}\\
    {\tt sceptical([neg(holds(running,3))])}\\
    {\tt sceptical([holds(running,6)])}\\
    {\tt sceptical([neg(holds(running,10))])}\\
    {\tt sceptical([holds(petrol,0)])}\\
    {\tt sceptical([holds(petrol,3)])}\\
    {\tt sceptical([holds(petrol,6)])}\\
    {\tt sceptical([neg(holds(petrol,10))])}
\end{center}
and fails on each converse query.\\


A more complex example (a variation of Example~\ref{VaccinationsExample})
that reasons with alternatives
from observations and ramifications is given as follows. Consider
the domain description $D_{i}$ given by:

\noindent
\\
\begin{tabular*}{\axiomwidth}{l@{\hspace*{\parindent}}l@{\extracolsep{\fill}}r@{}}
{} & {$\Expose \mbox{ {\tt initiates} } \Infected \mbox{ {\tt when} }
\{ \TypeA\}$} & {($D_{i}$1)}\\
{} & {$\Expose \mbox{ {\tt happens-at} } 3$} & {($D_{i}$2)}\\
{} & {$\neg \Infected \mbox{ {\tt holds-at} } 1$} & {($D_{i}$3)}\\
{} & {$\Infected \mbox{ {\tt holds-at} } 6$} & {($D_{i}$4)}\\
{} & {$\Expose \mbox{ {\tt initiates} } \Infected \mbox{ {\tt when} }
\{ \TypeB\}$} & {($D_{i}$5)}\\
{} & {$\InjectA \mbox{ {\tt initiates} } \Danger \mbox{ {\tt when} }
\{ \TypeA\}$} & {($D_{i}$6)}\\
{} & {$\InjectA \mbox{ {\tt initiates} } \Danger \mbox{ {\tt when} }
\{ \TypeB\}$} & {($D_{i}$7)}\\
{} & {$\InjectA \mbox{ {\tt happens-at} } 4$} & {($D_{i}$8)}\\
{} & {$\Allergic \mbox{ {\tt whenever} } \{\TypeA , \Infected \}$} &
{($D_{i}$9)}\\
{} & {$\Allergic \mbox{ {\tt whenever} } \{\TypeB\}$} & {($D_{i}$10)}\\
\end{tabular*}
\\

\noindent
From the observation at time $6$ the system
is able to reason (both backwards and forwards in time)
to prove that under any of the two possible
alternatives $\Danger$ holds after time $4$.
Thus, for any time {\tt t} after 4 the
system succeeds on
\begin{center}
    {\tt sceptical([holds(danger,t)])}.
\end{center}

\noindent
Similarly, the system succeeds on
\begin{center}
{\tt sceptical([holds(allergic,t)])},
 \end{center}
for any time {\tt t} after 3.

\section{Conclusions and Future Work}\label{ConclusionsSection}
We have described $\E$-RES, a Language $\E$ based system for reasoning about narratives
involving actions, change, observations and indirect effects via ramifications.
The functionality of $\E$-RES has been demonstrated both via theoretical
results and by testing with benchmark problems.
We have shown that the system is versatile enough to
handle a variety of reasoning tasks in simple domains. In particular, $\E$-RES can
correctly reason with incomplete information, both forwards and
backwards in time, from causes to effects and from effects to causes,
and about the indirect effects of action occurrences.

The system still needs to be tested with very large problems,
and possibly developed further to cope with the challenges that these pose.
In particular, the current handling of t-propositions and ramification statements
will probably be unsatisfactory for very large domains,
and techniques will need to be devised to select and reason with only the t- and r-propositions
that are {\em relevant} the the goal being asked.

Work is currently underway to extend the $\E$-RES system so that it can carry
out {\em planning}. The implementation will correspond to the $\E$-Planner
described in \cite{nmr00}. In our setting,  planning amounts to finding a suitable
set of h-propositions which, when added to the given domain description,
allow the entailment of a desired goal.
The $\E$-Planner is especially suitable for planning under incomplete information,
e.g. when
we do not have full knowledge of the
initial state of the problem,
and the missing
information cannot be ``filled in'' by performing additional actions,
(either because no actions exist which can affect the missing
information,
or because there is no time to perform such actions).
The planner needs to be able to reason correctly despite
this incompleteness, and construct plans (when such plans exist)
in cases where missing information is not necessary for achieving the desired goal.
For instance, in Example~\ref{VaccinationsExample}, if $(D_v3)$ and  $(D_v4)$
are missing, and the goal to achieve is \Protected\ {\tt holds-at} 4,
then the $\E$-Planner generates
$\InjectA \mbox{ {\tt happens-at} } T_1$ and
$\InjectB \mbox{ {\tt happens-at} } T_2$, with
$T_1, T_2 < 4$.


\subsection{Obtaining the System}\label{ObtainingSection}
Both $\E$-RES and codings of example test domains are available from the
Language $\E$ and $\E$-RES website at
\texttt{http://www.ucl.ac.uk/\~{{\ }}\hspace*{-0.3em}uczcrsm/LanguageE/}.


\bibliography{E-RES}
\bibliographystyle{aaai}

\end{document}